\documentclass[a4paper,12pt]{article}
\usepackage[centertags]{amsmath} 
\usepackage{amssymb}

\newcommand{\gpd}{GP$_\mathbb{X}$}
\newcommand{\card}{\mathop{\rm card}}
\newcounter{enumroman}
\newenvironment{romanitems}{\begin{list}{(\roman{enumroman})\hfill}{\usecounter{enumroman}
\setlength{\labelwidth}{\leftmargin}\addtolength{\labelwidth}{-1\labelsep}
\topsep=0mm plus 2pt\itemsep=0mm plus 1pt\parsep=0mm\itemindent=0mm}}{\end{list}}

\newtheorem{algorithm}{Algorithm}{\bf}{\rm}

\title{           Two Gaussian Approaches to Black-Box Optomization} 
\author{          Luk\'a\v{s} Bajer \and Martin Hole\v{n}a} 
\date{            }

\begin{document}
\bibliographystyle{plain}

\maketitle

\section{         CMA Evolution Strategy}
                  \emph{CMA-ES} \cite{hansen06cma,hansen01completely} is the
                  state-of-the-art evolutionary optimization method, at least in
                  the area of continuous black-box optimization. Basically, it
                  consists in generating new search points by sampling from a
                  multidimensional normal distribtion, the mean and variance of
                  which are updated from generation to generation. In
                  particular, the population
                  $x_1^{(g+1)},\dots,x_\lambda^{(g+1)}\in \mathbb{R}^d$ of the
                  $g+1$-st generation, $g\ge 1$, follows the normal distribution
                  with mean $m^{(g)}\in \mathbb{R}^d$ and variance
                  $(\sigma^{(g)})^2C^{(g)}\in \mathbb{R}^{d,d}$ resulting from
                  the update in the $g$-th generation,
\begin{gather} 
\label{cxd}        
                  x_i^{(g+1)}\sim N \left (m^{(g)},(\sigma^{(g)})^2C^{(g)}
                  \right ).
\end{gather}      
                  Here, $\sigma^{(g)}$ is the step size in the $g$-th
                  generation. $C^{(g)}\in \mathbb{R}^{d,d}$ and $\sigma^{(g)}>0$ are updated
                  separately, $C^{(g)}$ being in the most simple case obtained
                  as the unbiased empirical estimate based on the $g$-th generation:
\begin{gather}
                  C^{(g)}_\text{emp}=\frac{1}{ \lambda-1}\sum^\lambda_{i=1}
                  \left(x_i^{(g)}-\frac{1}{ \lambda }\sum^\lambda_{j=1}
                  x_j^{(g)} \right)\left(x_i^{(g)}-\frac{1}{ \lambda }\sum^\lambda_{j=1}
                  x_j^{(g)} \right)^\top.
\end{gather}      

                  The positive semidefinite matrix $C^{(g)}$ can be diagonalised
                  in the basis formed by its eigenvectors $b_1^{(g)},\dots,b_d^{(g)}$,
\begin{gather}
                  C^{(g)}=B^{(g)}D^{(g)}B^{(g)\top},
\end{gather}      
                  where $B^{(g)}= (b_1^{(g)} \cdots b_d^{(g)})$ and $D^{(g)}$ is
                  a diagonal matrix such that its $k$-th diagonal element is the
                  eigenvalue corresponding to the eigenvector
                  $b_k^{(g)}$. Consequently, the coordinates of points generated
                  according to (\ref{cxd}) in that basis,
                  $\{x_i^{(g+1)}\}^{B^{(g)}}$, are uncorrelated: 
\begin{gather}
                  \{x_i^{(g+1)}\}^{B^{(g)}}=B^{(g)\top}x_i^{(g+1)}\sim N \left
                  (B^{(g)\top}m^{(g)},(\sigma^{(g)})^2B^{(g)\top}C^{(g)}B^{(g)} \right).
\end{gather}

\section{         Optimization Based on Gaussian Processes} 
\label{gp}        
                  A \emph{Gaussian process (GP)} on a $d$-dimensional Euclidean
                  space $\mathbb{X}$ is a collection of random variables, \gpd$=
                  (f (x) )_{x\in\mathbb{X}}$, such that the joint distribution
                  of any finite number of them is a multidimensional normal
                  distribtion. Following \cite{buche05accelerating}, we
                  differentiate two ways of using GPs in black-box optimization: 
\begin{romanitems}
\item             As a \emph{surrogate model} to be optimized instead of the
                  black box objective function. On the found optimum or optima,
                  the original black-box objective function is then
                  evaluated. This use of GPs has been introduced in the popular
                  Efficient Global Optimization (EGO) algorithm
                  \cite{jones98efficient}. As the optimization method, also
                  evolutionary optimization can be used, in particular CMA-ES
                  \cite{buche05accelerating}, but this is not the only
                  possibility: For example, traditional low-degree polynomial
                  models, aka response surface models \cite{myers09response},
                  are much more efficiently optimized using traditional smooth
                  optimization methods.
\item             For the \emph{evolution control} of the evolutionary
                  optimization of the original objective function, i.e., for
                  controlling the composition of its population
                  \cite{elbeltagy01evolutionary,emmerich02metamodel,hoffmann06controlled,ulmer03evolution}.
\end{romanitems}   

                  Irrespective of the way in which a GP is used, its
                  construction is always based on a sequence
                  $(x_1,y_1),\dots,(x_n,y_n)\in\mathbb{X}\times\mathbb{R}$ of
                  training pairs and is subsequently employed to compute the
                  random variable $f (x)$ for $x\not\in\{x_1,\dots,x_n\}$. Since
                  the distribution of $(f (x),f (x_1),\dots,f (x_n))$ is
                  multidimensional normal, also the conditional distribution of
                  $f (x)$ conditioned on $f (x_1),\dots,f (x_n)$ is normal,
\begin{gather}
                  f (x)|f (x_1),\dots,f (x_n)\sim N (\mu (x;X,Y),\Sigma(x;X,Y) ),
\end{gather}      
                  where $X= (x_1,\dots,x_n),Y= (y_1,\dots,y_n),\mu ( \cdot
                  ;X,Y):\mathbb{X}\to\mathbb{X},\Sigma(\cdot
                  ;X,Y):\mathbb{X}\to\mathbb{X}^2$. There are two
                  commonly encountered possibilities how define the functions $\mu ( \cdot
                  ;X,Y)$, describing the conditional GP mean, and $\Sigma(\cdot ;X,Y)$,
                  describing the conditional GP variance:
\begin{enumerate}
\item             A GP is the \emph{superposition of a deterministic 
                  function $\bar{f}:\mathbb{X}\to\mathbb{R}$  and a GP with zero
                  mean}. For the latter, it can be shown
                  \cite{rasmussen06gaussian}  that the function describing its
                  mean fulfils  
\begin{gather} 
\label{gdm}        
                  (\forall x\in\mathbb{X})\; \mu_0 (x;X,Y)=K (x,X) (K
                  (X,X)+\sigma^2_{\text{noise}}I_n )^{-1}Y^\top ,
\end{gather}      
                  whereas the function describing its variance fulfills 
\begin{gather} 
\label{gds}        
                  (\forall x\in\mathbb{X})\;\Sigma_0(x;X,Y)=K (x,x)-K (x,X) (K
                  (X,X)+\sigma^2_{\text{noise}}I_n )^{-1}K (X,x).
\end{gather}      
                  Here, an i.i.d. Gaussian noise is assumed, $I_n$ is the
                  $n$-dimensional identity matrix,
                  $K:\mathbb{X}\times\mathbb{X}\to\mathbb{R}$ is a symmetric
                  function, and
\begin{multline}
                  K (x,X)= (K (x,x_1),\dots,K (x,x_n) ), K (X,x)=K (x,X)^\top,
\\
                  K (X,X)= (K (x_1,X)^\top,\dots,K (x_n,X)^\top)^\top. 
\end{multline}      
                  The resulting superposition then fulfils 
\begin{gather} 
\label{gnm}        
                  (\forall x\in\mathbb{X})\; \mu(x;X,Y)=\bar{f} (x)+K (x,X) (K
                  (X,X)+\sigma^2_{\text{noise}}I_n )^{-1} (Y-(\bar{f} (x_1),\dots,\bar{f}
                  (x_n)) )^\top, 
\end{gather}      
                  whereas $\Sigma=\Sigma_0$.
\item             A GP is the \emph{superposition of a Bayesian mean assuming the
                  multiplicative form $w\bar{f}$ and a GP with zero mean}, where $w$
                  is a random variable with $w\sim N (1,\sigma^2_w),\sigma_w>0, $
                  and $\bar{f}$ has the same meaning as above. In that case,
                  it can be shown \cite{rasmussen06gaussian} that for
                  $x\in\mathbb{X}$, 

\begin{multline}
\label{gbm}        
                  \mu(x;X,Y)=K (x,X) (K (X,X)+\sigma^2_{\text{noise}}I_n )^{-1}Y^\top+
\\
                  +(\bar{f} (x)- (\bar{f} (x_1),\dots,\bar{f} (x_n))(K
                  (X,X)+\sigma^2_{\text{noise}}I_n )^{-1}K (X,x))\hat{w},                  
\end{multline}    
\begin{multline}
\label{gbs}        
                  \Sigma(x;X,Y)=K (x,x)-K (x,X) (K
                  (X,X)+\sigma^2_{\text{noise}}I_n )^{-1}K
                  (X,x)+
\\
                  +\frac{\sigma^2_w(K (X,X)+\sigma^2_{\text{noise}}I_n
                  )^{-1}Y^\top+\hat{\kappa}^2}{1+\sigma^2_w(\bar{f}
                  (x_1),\dots,\bar{f} (x_n))(K (X,X)+\sigma^2_{\text{noise}}I_n
                  )^{-1}(\bar{f} (x_1),\dots,\bar{f} (x_n))^\top},
\end{multline}    
                  where 
\begin{gather}
                  \hat{w}=\frac{1+\sigma^2_w(\bar{f} (x_1),\dots,\bar{f} (x_n))(K
                  (X,X)+\sigma^2_{\text{noise}}I_n )^{-1}(\bar{f} (x_1),\dots,\bar{f}
                  (x_n))^\top}{1+\sigma^2_w(\bar{f} (x_1),\dots,\bar{f} (x_n))(K
                  (X,X)+\sigma^2_{\text{noise}}I_n )^{-1}Y^\top}, 
\\
                  \hat{\kappa}=(\bar{f} (x)- (\bar{f} (x_1),\dots,\bar{f} (x_n))(K
                  (X,X)+\sigma^2_{\text{noise}}I_n )^{-1}K
                  (X,x))
\end{gather}      
\end{enumerate}   

                  Simple, but frequently used examples of the function $K$
                  occurring in (\ref{gdm})--(\ref{gds}) and
                  (\ref{gbm})--(\ref{gbs}) include:
\begin{itemize}
\item             \emph{Squared exponential},
\begin{gather}
                  (\forall x,x'\in\mathbb{X})\; K
                  (x,x')=e^{-\frac{\|x-x'\|^2}{2\ell^2}} , \ell>0.  
\end{gather}      
                  Here, $\ell$ is a parameter called \emph{characteristic
                  length-scale}. It is a parameter of the function
                  $K_{\text{SE}}$, not of the Gaussian process, the process is
                  nonparametric. Therefore $\ell$ is referred to as a
                  \emph{hyperparameter}. 
\item             \emph{$\gamma$-Exponential},
\begin{gather} 
\label{gge}        
                  (\forall
                  x,x'\in\mathbb{X})\;K(x,x')=e^{-\left(\frac{\|x-x'\|}{\ell}
                  \right )^\gamma} , \ell>0,0<\gamma\le 2, 
\end{gather}      
                  which has 2 hyperparameters, $\lambda $ and $\gamma$.
\item             \emph{Dot product},
\begin{gather}
                  (\forall x,x'\in\mathbb{X})\;K(x,x')= (\sigma^2+x^\top
                  x')^p,\sigma\ge 0,p\in\mathbb{N},
\end{gather}      
                  which has 2 hyperparameters, $ \sigma $ and $p$, or in its
                  generalized version,
\begin{gather}
                  (\forall x,x'\in\mathbb{X})\;K(x,x')= (\sigma^2+x^\top\Sigma
                  x')^p,\Sigma\in\mathbb{R}^{d,d} \text{ positive definite},
\end{gather}      
                  which has, in addition, the matrix of hyperparameters
                  $\Sigma$, defining a dot product in general coordinates.
\end{itemize}

\subsection{      GP-based criteria to choose the points for evaluation} 
\label{gc}        
                  Whereas traditional response surface and surrogate models
                  employ basically only one criterion for the choice of points
                  in which the black box objective function should be evaluated,
                  namely the global or at least local minimum of the model (if
                  the optimization objective is minimization), GPs offer several
                  additional criteria:
\begin{romanitems}
\item             \emph{Minimum of a prescribed quantile (Q$_\alpha $)} of the
                  distribution of $f (x)|f (x_1),\dots,f (x_n)$, $\alpha \in
                  (0,1)$.  
\begin{gather} 
\label{eqq}        
                  x_\alpha^*=\arg\min_{x\in\mathbb{X}}q_\alpha  (N (\mu
                  (x;X,Y),\Sigma(x;X,Y) )).
\end{gather}      
                  Usually, (\ref{eqq}) is expressed using quantiles of the
                  standard normal distribution, $u_\alpha=q_\alpha (N (0,1) )$,
\begin{gather} 
\label{equ}        
                  x_\alpha^*=\arg\min_{x\in\mathbb{X}}\mu
                  (x;X,Y)+\sqrt{\Sigma(x;X,Y)}u_\alpha=\arg\min_{x\in\mathbb{X}}\mu
                  (x;X,Y)-\sqrt{\Sigma(x;X,Y)}u_{1-\alpha }.
\end{gather}      
                  For $\alpha=0.5$, (\ref{equ}) turns to the traditional global
                  minimum criterion, applied to  $\mu (\cdot ;X,Y)$. 
\item             \emph{Probability of improvement (PoI)}, 
\begin{gather}
                  x_\text{PI}^*=\arg\max_{x\in\mathbb{X}}P (f (x)< f_\text{min}|f
                  (x_1),\dots,f (x_n))=\phi \left(\frac{f_\text{min}-\mu
                  (x;X,Y)}{\sqrt{\Sigma(x;X,Y)}} \right),
\end{gather}     
                  where $\phi$ denotes the distribution function of $N (0,1)$,
                  $f_\text{min} $ is the minimum value found so far. More
                  generally, \emph{probability of improvement with respect to a
                  given} $T\le f_\text{min}$,
\begin{gather}
                  x_\text{PI|T}^*=\arg\max_{x\in\mathbb{X}}P (f (x)<T|f
                  (x_1),\dots,f (x_n))=\phi \left(\frac{T-\mu
                  (x;X,Y)}{\sqrt{\Sigma(x;X,Y)}} \right).
\end{gather}      
\item             \emph{Expected improvement (EI)},
\begin{multline} 
\label{eei}        
                  x_\text{EI}^*=\arg\max_{x\in\mathbb{X}}E ( (f_\text{min}-f
                  (x))I (f(x)<f_\text{min} ) |f (x_1),\dots,f (x_n)),
\\
                  \text{where } I (f(x)<f_\text{min} )=
\begin{cases}
                  1 & f(x)<f_\text{min},
\\
                  0 & f(x)\ge f_\text{min}.
\end{cases}      
\end{multline}      
                  Introducing the normalized mean improvement
                  $\nu:\mathbb{X}\to\mathbb{R}$, 
\begin{gather}
                  (\forall x\in\mathbb{X})\;\nu (x)=\frac{f_\text{min}-\mu
                  (x;X,Y)}{\sqrt{\Sigma(x;X,Y)}},
\end{gather}      
                  and the notation $ \varphi $ for the density of $N (0,1)$,
                  (\ref{eei}) can be expressed as \cite{jones01taxonomy}
\begin{gather}
                  x_\text{EI}^*=\arg\max_{x\in\mathbb{X}}\sqrt{\Sigma(x;X,Y)}
                  (\nu (x)\phi (\nu (x) )+\varphi (\nu (x) ) ) ). 
\end{gather}      
\end{romanitems}

\section{         Possible Synergy}
                  Directly connecting both considered Gaussian approaches is not
                  possible because the normal distribution in CMA-ES is a
                  distribution on the input space of the objective function
                  (fitness), whereas the normal distribution in GPs is on the
                  space of its function values. Nevertheless, it is still
                  possible to achieve some synergy through using information
                  from CMA-ES for the GP, and/or using information from the GP
                  for CMA-ES. According to whoat was recalled at the beginning
                  of Section~\ref{gp}, the latter possibility corresponds to
                  using GP for the evolution control of CMA-ES.

\subsection{      Using Information from CMA-ES for the GP}
                  We see 2 straightforward possibilities where some
                  information from CMA-ES can be used in the GP.
\begin{enumerate}
\item             The function $\bar{f}$ occurring in (\ref{gnm}), (\ref{gbm})
                  and (\ref{gbs}) can be constructed using the fitness values
                  $f(x_1^{(g)}),\dots$ $\dots,f(x_\lambda^{(g)})$ of individuals from
                  some particular generation or several generations of
                  CMA-ES. Its construction can be as simple as setting $\bar{f}$
                  to a constant aggregating the considered fitness values, e.g.,
                  their mean or weighted mean, but it can also consist in
                  training, with those values, a response surface model,
                  principally of any kind.
\item             Kruisselbrink et~al. \cite{kruisselbrink10robust}, who combine
                  CMA-ES with a GP using the $ \gamma$-exponential function $K$,
                  propose to employ in (\ref{gge}) the Mahalanobis distance of
                  vectors $x$ and $x'$ given by the covariance matrix obtained
                  in the $g$-th generation of CMA-ES, $(\sigma^{(g)})^2C^{(g)}$,
                  instead of their Euclidean distance. This is actually a
                  specific consequence of another possibility of using
                  information from CMA-ES for the GP -- replacing the original
                  space of $d$-dimensional vectors, $\mathbb{R}^d$, by the space
                  of their principal components with respect to $C^{(g)}$. In
                  this context, it is worth recalling that in
                  \cite{bouzarkouna10investigating,bouzarkouna11local,kern06local},
                  Mahalanobis instead of Euclidean distance was used when
                  combining CMA-ES with quadratic response surface models in
                  \cite{bouzarkouna10investigating,bouzarkouna11local,kern06local}. In
                  the approach presented there, the space of the principal
                  components with respect to $C^{(g)}$ is used, together with an
                  estimate of density, to locally weight the model predictions
                  with respect to the considered input. In this way, a
                  connection to the other kind of synergy is established, i.e.,
                  to the evolution control of CMA-ES, giving us the possibility
                  to use both kinds in combination.
\end{enumerate}

\subsection{      GP-Based Evolution Control of CMA-ES}
                  We intend to test the following approaches to using GP for the
                  evolution control of CMA-ES. 
\begin{figure}[bt]
\centering 
\begin{minipage}[t]{.99\linewidth}
\begin{algorithm} 
                  \mbox{} 
\\
                  \emph{Input:} generation $g$, points
                  $\tilde{x}_1,\dots,\tilde{x}_{\lambda'}\in \mathbb{R}^d\sim N
                  (m^{(g)},(\sigma^{(g)})^2C^{(g)})$, their linear ordering
                  $\prec_c,c\in\{\text{Q}_\alpha,\text{PoI},\text{EI}\}$ with
                  $\alpha \in (0,1)$ according to a GP-based criterion to choose
                  the points for evaluation, $\lambda \in
                  \mathbb{N},\lambda<\card
                  \{\tilde{x}_1,\dots,\tilde{x}_{\lambda'}\},
                  k\in\{1,\dots,\lambda \} $.
\\
                  \emph{Step 1.} Perform $k$-means clustering of
                  $S=\{\tilde{x}_1,\dots,\tilde{x}_{\lambda'}\},
                  k\in\{1,\dots,\lambda \}$ into sets $S_1,\dots,S_k$. 
\\
                  \emph{Step 2.} For $j=1,\dots,k$, choose
                  $x_j^{(g+1)}=\max_{\prec_c}S_j$.
\\
                  \emph{Step 3.} For $j=k+1,\dots,\lambda $, choose
                  $x_j^{(g+1)}=\max_{\prec_c}S \setminus \{x_i^{(g+1)}:i=1,j-1\}$. 
\\
                  \emph{Output:} Points $x_1^{(g+1)},\dots,x_\lambda^{(g+1)}$ to
                  be evaluated by the original fitness function.
\end{algorithm}   
                  \caption{Algorithm of the proposed strategy for choosing the $
                  \lambda$ points to be evaluated by the original fitness from
                  among the $ \lambda' $ evaluated by the Gaussian process\label{a1}}
\end{minipage}   
\end{figure} 
\begin{romanitems}
\item             \emph{Basic approach.} In the $(g+1)$-th generation, $
                  \lambda'$ points $\tilde{x}_1,\dots,\tilde{x}_{\lambda'}\in
                  \mathbb{R}^d$ are sampled from the distribution $N
                  (m^{(g)},(\sigma^{(g)})^2C^{(g)})$, where $ \lambda'$ is
                  several to many times larger then $ \lambda $. For all of
                  them, a selected criterion from among those introduced in
                  \ref{gc} is computed. Based on the value of that criterion,
                  the $ \lambda $ points $x_1^{(g+1)},\dots,x_\lambda^{(g+1)}$
                  for the evaluation by the original black-box fitness are
                  chosen. This can be done according to various strategies, we
                  intend to use the one described in Algorithm~\ref{a1}, with
                  which we have a good experience from using radial basis
                  function networks for the evolution control in the
                  evolutionary optimization of catalytic materials
                  \cite{12philadelphia}. The linear ordering $\prec_c$ stands in
                  the cases $c=$PoI and $c=$EI for $\le$, in the case 
                  $c=\text{Q}_\alpha $ for $\ge$. The use of a linear ordering
                  relates the proposed approach to ranking-based evolution control of
                  CMA-ES
                  \cite{bouzarkouna10investigating,bouzarkouna11local,kern06local,ulmer03evolution},
                  as well as to surrogate modelling of CMA-ES by ordinal
                  regression
                  \cite{loshchilov10comparison,loshchilov12self,loshchilov13intensive,runarsson06ordinal}. Most
                  similar is the approach by Ulmer
                  et~al. \cite{ulmer03evolution}, the difference being that they
                  don't use clustering.
\item             \emph{GP on low-dimensional projections} attempts to improve
                  the basic approach in view of the experience reported in the
                  literature \cite{kern06local} and obtained also in our earlier
                  experiments \cite{bajer13model} that GPs are actually advantageous
                  only in low dimensional spaces. Instead of the sampled points
                  $\tilde{x}_1,\dots,\tilde{x}_{\lambda'}$, the GP is
                  trained only with their first $\ell<d$ principal components
                  with respect to $C^{(g)}$,
                  $\tilde{x}_1^{[\ell]},\dots,\tilde{x}_{\lambda'}^{[\ell]}$,
                  i.e., with the projections of
                  $\tilde{x}_1,\dots,\tilde{x}_{\lambda'}$ to the
                  $\ell$-dimensional space $\mathbb{X}_\ell=\text{span}
                  (b_1^{(g)},\dots,b_\ell^{(g)})$, provided the orthonormal
                  eigenvectors $b_i^{(g)}$ are enumerated according to
                  decreasing eigenvalues. 
\item             \emph{GP on low-dimensional projections within restricted
                  distance} attempts to decreaase the deterioration of the GP
                  on low-dimensional projections with respect to the GP on the
                  original sampled points
                  $\tilde{x}_1,\dots,\tilde{x}_{\lambda'}$ by using only points
                  $\tilde{x}_i$ within a prescribed small distance $\epsilon$
                  from their respective projections $\tilde{x}_i^{[\ell]}$. To
                  this end, points from the distribution $N
                  (m^{(g)},(\sigma^{(g)})^2C^{(g)})$ are resampled until $
                  \lambda'$ points $\tilde{x}_i$ are obtained fulfilling 
\begin{gather}
                  \|\tilde{x}_i-\tilde{x}_1^{[\ell]}\|<\epsilon\text{, or
                  equivalently, }
                  \|\tilde{x}_i\|^2-\|\tilde{x}_1^{[\ell]}\|^2<\epsilon^2. 
\end{gather}      
\item             \emph{Two-stage sampling.} Trainig the GP for the $(g+1)$-th
                  generation can easily suffer
                  from the lack of training data in a part of the search
                  space where a substantial proportion of the
                  points  $\tilde{x}_1,\dots,\tilde{x}_{\lambda'}$ will be
                  sampled. To alleviate it, the  points
                  $x_1^{(g+1)},\dots,x_\lambda^{(g+1)}$ to be evaluated by the
                  original fitness can be sampled in two stages:
\begin{enumerate}
\item             First, points $x_1^{(g+1)},\dots,x_{\lambda''}^{(g+1)}$, where
                  $\lambda''<\lambda $, are sampled from the distribution $N
                  (m^{(g)},(\sigma^{(g)})^2C^{(g)})$, evaluated by the original
                  fitness, and included into training the GP.
\item             Then, $\lambda'$ points
                  $\tilde{x}_1,\dots,\tilde{x}_{\lambda'}$ are sampled from the
                  same distribution, and for them, a selected criterion from
                  among those introduced in \ref{gc} is computed, based on
                  which the points $x_{\lambda''+1}^{(g+1)},\dots,x_\lambda^{(g+1)}$
                  for the evaluation by the original black-box fitness are
                  chosen. To this end, again  Algorithm~\ref{a1} can be used,
                  with the following changes:
\begin{itemize}
\item             In the input, the numbers $\lambda'$ and $k$ have now to
                  fulfill 
\begin{gather}
                  \lambda-\lambda''<\card
                  \{\tilde{x}_1,\dots,\tilde{x}_{\lambda'}\},
                  k\in\{1,\dots,\lambda-\lambda'' \}. 
\end{gather}      
\item             In Step 3, $x_j^{(g+1)}=\max_{\prec_c}S \setminus
                  \{x_i^{(g+1)}:i=1,j-1\}$ is chosen only for
                  $j=k+1,\dots,\lambda-\lambda'' $.
\item             The output contains only the points
                  $x_{\lambda''+1}^{(g+1)},\dots,x_\lambda^{(g+1)}$. 
\end{itemize}     
\end{enumerate}   
                  Needless to say, this approach has to be combined with some of
                  the approaches (i)--(iii) and can be combined with any of
                  them.
\item             \emph{Generation-based evolution control.} Whereas the
                  previous four approaches represent, in terms of
                  \cite{jin05comprehensive,jin01managing},
                  individual-based evolution control, we want to test also one
                  approach that is generation-based, in the sense that the
                  desired number $\lambda $ of points is evaluated by the
                  original black-box fitness function only in selected
                  generations. Similarly to the evolution strategy in
                  \cite{loshchilov12self,loshchilov13intensive}, our approach
                  selects those generations adaptively, according to the
                  agreement between the ranking of considered points by the
                  surrogate model and by the black-box fitness. Differently to
                  that strategy, however, we want to base the estimation if that
                  agreement not on the generations in which $\lambda $ points
                  have been evaluated by the black-box fitness, but on
                  evaluating a small and evolvable number $\lambda'''$ of
                  additional points in each generation. In this context, it is
                  important that in situations when the fitness is evaluated
                  empirically, using some measurement or testing, the evaluation
                  hardware causes the evaluation costs to increase step-wise,
                  and to remain subsequently constant for some
                  $\lambda_\text{hw}$ evaluated points (e.g. in the optimization
                  of catalyst preformance described in \cite{12philadelphia},
                  $\lambda_\text{hw}$ is the number of channels in the chemical
                  reactor in which the catalysts are tested). In such a
                  situation, the costs of evaluation are the lowest if $\lambda
                  $ is a multiple of $\lambda_\text{hw}$, and if the evaluation
                  of the $\lambda'''$ additional points in each generation is
                  cumulated for $n_\text{hw}$ generations such that
\begin{gather}
                  n_\text{hw}=\max_{n\in \textbf{N}}n\lambda'''\le\lambda_\text{hw}.
\end{gather}      Setting $n_\text{hw}=1$ covers the case when such a situation
                  does not occur and the evaluation costs increase linearly with
                  the number of points. If $g_\text{last} $ is the last
                  generation in which $\lambda $ points have been evaluated by
                  the black-box fitness, then for
                  $g=g_\text{last},\dots,g_\text{last}+n_\text{hw}-1$, points
                  $\tilde{x}_1^{g+1},\dots,\tilde{x}_{\lambda'}^{g+1}\in
                  \mathbb{R}^d$ are sampled from the distribution $N
                  (m^{(g)},(\sigma^{(g)})^2C^{(g)})$, from which the points
                  $x_1^{(g+1)},\dots,x_{\lambda'''}^{(g+1)}$ are selected by
                  Algorithm~1, in which the input $\lambda$ is replaced with
                  $\lambda'''$. Subsequently, all the points
                  $x_1^{(g_\text{last}+1)},\dots,x_{\lambda'''}^{(g_\text{last}+n_\text{hw})}$
                  are evaluated by the black-box fitness, and for each
                  generation $g=g_\text{last}+1,\dots,g_\text{last}+n_\text{hw}$,
                  the agreement between the ranking of
                  $x_1^g,\dots,x_{\lambda'''}^g$ by the current Gaussian
                  process, GP$_\text{current}$, and the black-box fitness is
                  estimated. If the agreement is sufficient in all
                  considered generations, then the procedure is repeated using
                  $g_\text{last}+n_\text{hw}$ instead of $g_\text{last}$ and
                  unchanged $\lambda'''$. Otherwise, additional points
                  $x_{\lambda'''+1}^g,\dots,x_{\lambda'''+\lambda }^g$ are
                  selected from $\tilde{x}_1^g,\dots,\tilde{x}_{\lambda'}^g$ for
                  the first generation $g$ in which the agreement was not
                  sufficient. Then a new Gaussian process, GP$_\text{new}$, is
                  trained using the training set 
\begin{gather}
                  T_\text{new}
                  =\bigcup_{g'=1}^g\{x_1^{g'},\dots,x_{\lambda'''}^{g'}\}\cup
                  \{x_{\lambda'''+1}^g,\dots,x_{\lambda'''+\lambda }^g\},  
\end{gather}      
                  optionally including also some or all points from the set
                  $T_\text{current}$ used for training GP$_\text{current}$. At that occasion, aslo the value of $\lambda'''$ can
                  be changed.

                  Provided the conditions $1\le\lambda'''< \lambda $ and
                  $\lambda'''+\lambda\le\lambda'$ are fulfilled, the value of
                  $\lambda'''$ can be arbitrary. Needless to say, the smaller
                  $\lambda'''$, the larger will be the proportion of points from
                  the generation in which the GP was trained in its training
                  set, and the more similar the obtained GP will normally be to
                  a GP trained only with data from that generation, which is the
                  usual way of using surrogate models in traditional
                  generation-based strategies
                  \cite{jin05comprehensive,jin01managing,loshchilov12self,loshchilov13intensive}. It
                  is also worth pointing out that our generation-based evolution
                  control was explained here as a counterpart to the basic
                  individual-based approach (i), but counterparts to the
                  approaches (ii) and (iii) are possible as well.
\end{romanitems}

\bibliography{odkazy} 
\end{document}